\documentclass[letterpaper]{article} 
\usepackage{aaai25}  
\usepackage{times}  
\usepackage{helvet}  
\usepackage{courier}  
\usepackage[hyphens]{url}  
\usepackage{graphicx} 
\usepackage{natbib}  
\usepackage{caption} 
\usepackage{algorithm}
\usepackage{algorithmic}

\usepackage{newfloat}
\usepackage{listings}
\DeclareCaptionStyle{ruled}{labelfont=normalfont,labelsep=colon,strut=off} 
\floatstyle{ruled}
\newfloat{listing}{tb}{lst}{}
\floatname{listing}{Listing}
\usepackage{arydshln}

\title{Just What You Desire: Constrained Timeline Summarization with Self-Reflection for Enhanced Relevance}
\author{
    Muhammad Reza Qorib\textsuperscript{\rm 1},
    Qisheng Hu\textsuperscript{\rm 2}\thanks{Work done while Qisheng Hu was affiliated with the National University of Singapore.},
    Hwee Tou Ng\textsuperscript{\rm 1}
}
\affiliations{
    \textsuperscript{\rm 1}Department of Computer Science, National University of Singapore\\
    \textsuperscript{\rm 2}College of Computing and Data Science, Nanyang Technological University \\

    mrqorib@u.nus.edu, qisheng001@e.ntu.edu.sg, nght@comp.nus.edu.sg
}

\begin{document}

\maketitle

\begin{abstract}
Given news articles about an entity, such as a public figure or organization, timeline summarization (TLS) involves generating a timeline that summarizes the key events about the entity. However, the TLS task is too underspecified, since what is of interest to each reader may vary, and hence there is not a single ideal or optimal timeline. In this paper, we introduce a novel task, called Constrained Timeline Summarization (CTLS), where a timeline is generated in which all events in the timeline meet some constraint. An example of a constrained timeline concerns the legal battles of Tiger Woods, where only events related to his legal problems are selected to appear in the timeline. We collected a new human-verified dataset of constrained timelines involving 47 entities and 5 constraints per entity. We propose an approach that employs a large language model (LLM) to summarize news articles according to a specified constraint and cluster them to identify key events to include in a constrained timeline. In addition, we propose a novel self-reflection method during summary generation, demonstrating that this approach successfully leads to improved performance.
\end{abstract}

\begin{links}
    \link{Code and Data}{https://github.com/nusnlp/reacts}
\end{links}

\section{Introduction}
In today's internet era, the rapid and massive flow of information makes it hard to stay updated, particularly for topics with extensive coverage over time. In the United States alone, \citet{Hamborg2018BiasawareNA} report that more than 5,000 news articles are being published every day. To help readers quickly grasp important information, many news platforms offer news in a timeline format, especially for important topics that progress over time, such as pandemics\footnote{\url{https://www.cdc.gov/museum/timeline/covid19.html}} or conflicts\footnote{\url{https://www.usnews.com/news/best-countries/slideshows/a-timeline-of-the-russia-ukraine-conflict}}.

\begin{table}[!ht]
\centering
\setlength{\tabcolsep}{1mm}
\begin{tabular}{p{0.21\linewidth} p{0.72\linewidth}}
    \hline
    \multicolumn{2}{l}{\textbf{Original Timeline}} \\
    \hline
    2003-11-19: & Receives the National Book Foundation Medal for Distinguished Contribution to American Letters. \\
    2015-09-10: & Is awarded the National Medal of Arts by US President Barack Obama. \\
    2015-11-03: & Releases a collection of short stories entitled “The Bazaar of Bad Dreams.” \\
    2018-07-25: & King is an executive producer of a show written for the streaming service Hulu. The series, “Castle Rock,” is named after the fictional small Maine town that provides the setting for various King books and stories. \\
    2020-04-21: & King’s latest book, “If It Bleeds” is published. The book is a compilation of four novellas. \\
    2021-06-04: & The miniseries “Lisey’s Story,” adapted by King and based on his 2006 novel of the same name, premieres on Apple TV+. \\
    2022-09-06: & King’s novel “Fairy Tale” is published. \\
    \hline
    \multicolumn{2}{l}{} \\
    \hline
    \multicolumn{2}{l}{\underline{\textbf{Constrained Timeline}}} \\
    \multicolumn{2}{l}{Constraint: Focus on Stephen King's book releases.} \\
    \hline
    2015-11-03: & Releases a collection of short stories entitled “The Bazaar of Bad Dreams.” \\
    2020-04-21: & King’s latest book, “If It Bleeds” is published. The book is a compilation of four novellas. \\
    2022-09-06: & King’s novel “Fairy Tale” is published. \\
    \hline
\end{tabular}
\caption{An unconstrained timeline of Stephen King and a constrained version focusing on Stephen King's book releases.}
\label{tab:example}
\end{table}

The task of summarizing news articles, or any collections of text documents, into timelines is called timeline summarization (TLS). Timeline summarization aims to automatically condense long-running news topics into temporally ordered time-stamped textual summaries of events on a particular topic. Timeline summarization aims to include any important events into the timeline without considering particular aspects that the readers are interested in.

To take a reader's interest into account when generating a timeline, we propose a new task called constrained timeline summarization (CTLS). Constrained timeline summarization offers personalization that TLS lacks. For example, a reader may want to automatically retrieve the timeline of Stephen King's book publication (Table \ref{tab:example}). In this example, Stephen King's national awards are irrelevant to the reader even though they are generally considered important events in Stephen King's life.

The contributions of this paper are as follows:
\begin{itemize}
    \item We propose a new task, constrained timeline summarization. The task has real-life applications.
    \item We present a new test set to benchmark models on the constrained timeline summarization task.
    \item We present an effective method that utilizes large language models without any need for training or fine-tuning.
    \item We propose a novel self-reflection method to produce a more relevant constrained event summary and demonstrate that self-reflection helps in generating more relevant constrained timelines.
\end{itemize}

\section{Related Work}
In this section, we briefly discuss related work on timeline summarization, query-based summarization, and update summarization. Constrained timeline summarization can be viewed as an amalgamation of the first two tasks.

\subsection{Timeline Summarization}
Previous work on timeline summarization can be categorized into three main approaches: direct summarization, date-wise approaches, and event detection approaches.

\subsubsection{Direct Summarization}
In this approach, a collection of documents is treated as a set of sentences to be directly extracted. Sentence extraction can be performed by optimizing sentence combinations \cite{martschat-markert-2018-temporally} or by ranking sentences \cite{10.1145/1008992.1009065}. This category also includes methods that treat the task as an extension of multi-document summarization, where the goal is to generate a summary from multiple documents \cite{10.1145/383952.383954, yu-etal-2021-multi}.

\subsubsection{Date-wise Approach}
In this approach, the task is divided into two steps: identifying important dates and summarizing events that occurred on those dates. Most methods employ supervised techniques to select the dates. For instance, \citet{gholipour-ghalandari-ifrim-2020-examining} propose a classification or regression model to predict date importance, while \citet{tran-etal-2015-joint} utilize graph-based ranking for date selection.

\subsubsection{Event Detection}
In this approach, the system first detects important events from the articles by clustering them based on similarity. It then ranks and selects the most important clusters and summarizes them into event descriptions. Various techniques have been proposed for clustering, including Markov clustering on bag-of-words features \cite{gholipour-ghalandari-ifrim-2020-examining}, dynamic affinity-preserving random walks \cite{Duan2020ComparativeTS}, event graph compression \cite{li-etal-2021-timeline}, date graph model \cite{10.1145/3404835.3462954}, heterogeneous graph attention networks \cite{you-etal-2022-joint}, and even large language models \cite{hu-etal-2024-moments}.

\subsection{Query-Based Summarization}
Query-based summarization, also called query-focused, topic-based, or user-focused summarization, aims to extract and summarize information that users are specifically interested in from a large number of texts. Essentially, it is a type of summarization that leverages user-provided query information.

Early approaches to query-based summarization mainly score or rank the relevance of each sentence in the document to the query based on predefined features \cite{7377323}. Sentences with the highest scores are then extracted to create the summary. Relevance scoring can be performed in an unsupervised manner by utilizing lexical and semantic features \cite{conroy-etal-2006-topic, 10.5120/11587-6925} or in a supervised manner by training regressor models \cite{Mani1998MachineLO, OUYANG2011227}. Document graphs are also often employed when dealing with multiple documents \cite{4042277, WANG2013271}.

Due to the effectiveness of transformers, recent query-based summarization methods are predominantly based on transformer models, including large language models. For example, \citet{Laskar2020QueryFA} incorporate query relevance into BERTSUM \cite{liu-lapata-2019-text}, while \citet{10.1145/3477495.3531901} integrate a query-attentive semantic graph with sequence-to-sequence transformers. Fine-tuning large language models has also been explored, such as in the work by \citet{xu-etal-2023-lmgqs}, who fine-tune BART \cite{lewis-etal-2020-bart}, and \citet{cao2024idealleveraginginfinitedynamic}, who fine-tune Llama 2 \cite{touvron2023llama2openfoundation} using custom adapters.

\subsection{Update Summarization}
Update summarization is the task of generating a short summary from a set of documents $A$ under the assumption that users have read a set of documents $B$ \cite{tac08}. Update summarization has a different objective from timeline summarization, but the methods proposed for it often bear some resemblance to the event detection approach for timeline summarization, notably in determining the novelty of the information from set $A$ in relation to set $B$ \cite{10.1145/1600193.1600239}. In the context of timeline summarization, novelty detection involves determining whether the extracted events from set $A$ are new events that are not present in set $B$.

\begin{figure*}[t]
\centering
\includegraphics[width=0.95\textwidth]{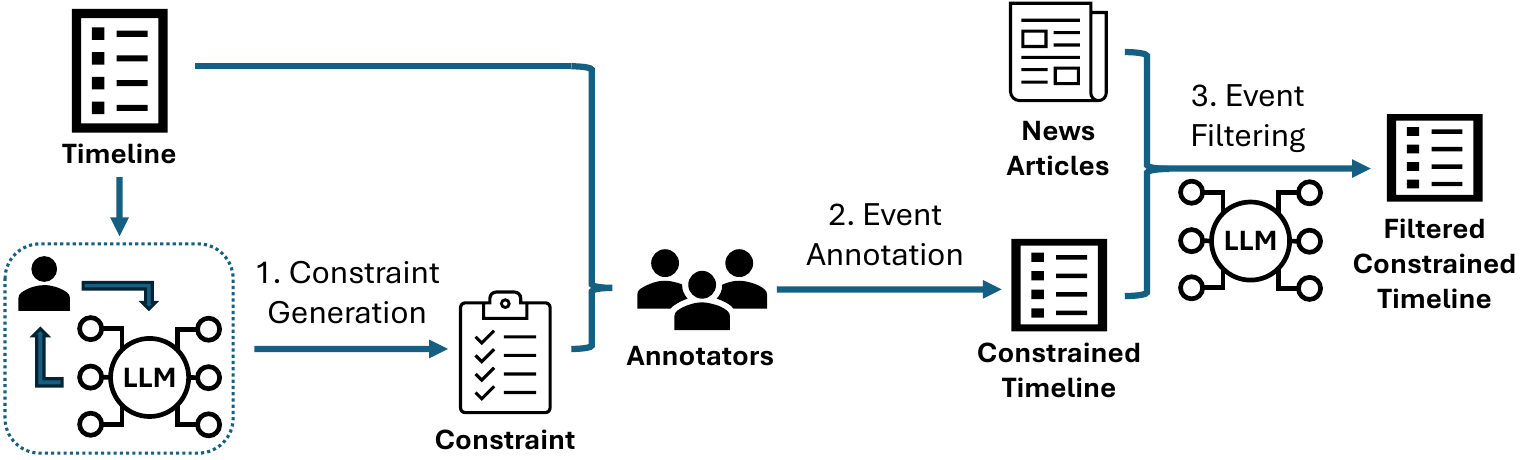} 
\caption{The dataset creation process consists of three steps: constraint generation, event annotation, and event filtering. Human annotators are tasked with determining whether an event in a timeline adheres to a constraint or not. The list of events that adheres to a constraint becomes a constrained timeline.}
\label{fig:dataset_creation}
\end{figure*}

\section{Dataset}
To benchmark the constrained timeline summarization task, we propose a novel test set called \textsc{CREST} (\underline{C}onstraint \underline{R}estrictions on \underline{E}ntities to \underline{S}ubset \underline{T}imelines). \textsc{CREST} consists of 235 timelines from 47 public figures or institutions (entities). We derive these timelines from the \textsc{ENTITIES} dataset \cite{gholipour-ghalandari-ifrim-2020-examining}, which were crawled from CNN Fast Facts. For each entity, we generate 5 pairs of constraints and corresponding subset timelines. The article pool is sourced from the \textsc{ENTITIES} dataset, which was collected from The Guardian using the official API\footnote{\url{https://open-platform.theguardian.com/}}. As such, our dataset is limited to British and American news sources. The dataset creation process involves constraint generation, event annotation, and event filtering.

\subsection{Constraint Generation}
We generate constraints by prompting GPT-4o\footnote{\texttt{gpt-4o-2024-08-06}} and manually selecting the best 5 constraints for each timeline. The prompt instructs GPT-4o to propose single-sentence constraints in the format ``Focus on ...''. To ensure that we have a variety of constraints, we query GPT-4o with four different types of prompts: general, numerical, relational, and geographical. The general prompt asks GPT-4o for constraint suggestions without any additional specification. The numerical prompt asks GPT-4o to generate constraints that contain ordinal phrases (e.g., first, second, etc.) or time indicators (e.g., timestamp, month, year, etc.). The relational prompt focuses on constraints involving some relationship between the entity and other public figures or institutions (e.g., ``Focus on Stephen King's interactions with President Barack Obama.''), while the geographical prompt asks for constraints with geographical information.

We find that GPT-4o generally suggests good constraints, but occasionally, the suggestions include hallucinations (e.g., constraints referencing non-existent events in the timeline) or are overly specific (e.g., only applicable to one event). Therefore, a human-in-the-loop process is essential to curate a good set of constraints for each timeline. Human intervention involves selecting the proposed constraints and modifying them to be more general.

\subsection{Event Annotation}
Given a list of events $E_t$ from timeline $t$ and a set of constraints $C_t$ for timeline $t$, we build the constrained timelines by asking human annotators to label whether each event in the timeline adheres to each constraint. All constraints are applied to all events in the timeline, resulting in $|E_t| \times |C_t|$ assertions. Each annotator is provided with the complete timeline (containing all events) and the full set of constraints for that timeline.

We recruited 4 university students with strong English proficiency as annotators. To ensure high-quality annotations, the annotators completed a qualifying test by performing annotations on a different timeline. The annotators performed the task for approximately four hours and were compensated above standard rates (\$22.65/hour). We found that our test set had high inter-annotator agreement, with an exact match percentage of 94.7\% and a Cohen’s kappa of 0.78 between the first and second annotators and an exact match percentage of 96.2\% and a Cohen's kappa of 0.88 between the third and fourth annotators.

\subsection{Event Filtering}
One challenge with the \textsc{ENTITIES} dataset is that the article pool and timelines were collected independently from different sources. This causes a mismatch between the events covered by the articles and those included in the timelines. As a result, some important events in the ground-truth timelines are not covered by the article pool, making it impossible for the model to generate them without external knowledge. In such cases, even a human would not be able to achieve a perfect score.

To avoid unfairly penalizing an automatically constructed model, we provide an additional evaluation setting in which events in the timelines that are not covered by the article pool are filtered out. Given that the article pool contains more than forty-five thousand news articles, manually checking event coverage would be too costly and labor intensive. Following previous work \cite{Gilardi2023ChatGPTOC}, we utilize GPT-4o to check each article for information related to the events in question.

It is important to note that we only filter out events from the timelines, while the article pool remains unchanged. We assume that the ground truth timelines are comprehensive lists of all significant events related to the entities. We report the statistics of our dataset for both the full and filtered settings in Table \ref{tab:statistics}.

\begin{table}[t]
\centering
\begin{tabular}{l|cc}
    \hline
    Statistics & All Events & Filtered \\
    \hline
    \# topics & 47 & 47 \\
    \# timelines & 233 & 201 \\
    \# events & 1031 & 667 \\
    Avg. \# articles per topic & 959 & 959 \\
    Avg. \# timelines per topic & 4.96 & 4.28 \\
    Avg. \# events per timeline & 4.42 & 3.32 \\
    \hline
\end{tabular}
\caption{Statistics of our proposed dataset (\textsc{CREST}).}
\label{tab:statistics}
\end{table}

\begin{figure*}[t]
\centering
\includegraphics[width=0.95\textwidth]{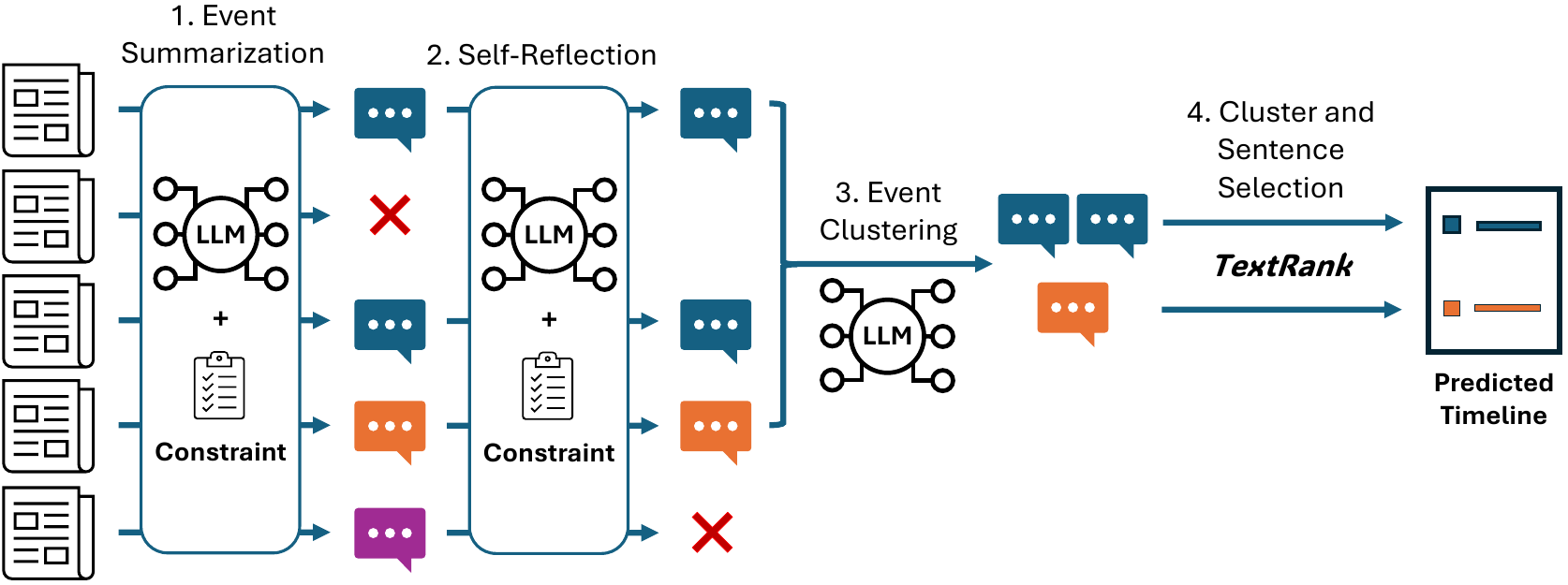}
\caption{Illustration of our method for constrained timeline summarization (\textsc{REACTS}). The method consists of four steps: event summarization, self-reflection, event clustering, and cluster and sentence selection.}
\label{fig:method}
\end{figure*}

\begin{algorithm}
\caption{Method}\label{alg:method}
\begin{algorithmic}
\REQUIRE A queue of articles $A$, a topic keyword $q$, a constraint $c$, a new article $a_i$ that arrived at time $i$, the event database $D$, the event clusters $G$, the retrieval limit $N$, the number of dates $l$ in the timeline, the number of sentences per date $k$ in the timeline.
\ENSURE A timeline $T_{q, c}$ about topic $q$ following the constraint $c$, comprising $l$ timestamped event descriptions, each with $k$ sentences.
\STATE $a_i \gets \textsc{Dequeue}(A)$
\STATE $e_i \gets \textsc{ConstrainedTopicSum}(a_i, q, c)$ 
\IF{$e_i \neq $ \texttt{NULL}}
    \IF{\textsc{AdhereToConstraint}$(e_i, q, c)$}
        \STATE $Edges \gets \{\}$
        \FORALL{$e_j \in \textsc{Retrieve}(D, e_i, N)$}
            \IF{$\textsc{SameEvent}(e_i, e_j)$}
                \STATE $Edges \gets Edges \cup \{[e_i, e_j]\}$
            \ENDIF
        \ENDFOR
        \STATE $G \gets \textsc{UpdateClusters}(G, Edges)$
        \STATE $D \gets \textsc{Insert}(D, e_i)$
    \ENDIF
\ENDIF
\STATE $Clusters \gets \textsc{RankClusters}(G, l)$
\STATE $T_{q, c} \gets []$
\STATE $j \gets 1$
\FORALL{$v \in \textsc{SortByTime}(Clusters)$}
    \STATE $T_{q,c}[j] \gets \textsc{Summarize}(v, k)$
    \STATE $j \gets j + 1$
\ENDFOR
\RETURN $T_{q, c}$
\end{algorithmic}
\end{algorithm}

\section{Problem Definition}
Constrained timeline summarization is a task to generate a timeline $T$ that includes important events related to a topic and adhering to a constraint, given a list of input documents. The input comprises temporally ordered documents $A = \{a_1, a_2, ...\}$ related to a specific topic $q$, a constraint $c$, the expected number of dates $l$ in the timeline, and the expected number of sentences per date $k$. The system-generated timeline $T$ will be evaluated against a ground-truth timeline $R$. Similar to most timeline summarization datasets, the list of documents in this dataset is a list of chronologically ordered news articles. The constraint is a natural language sentence that specifies the kind of events related to the topic $q$ that should be included in the timeline.

\section{Method}
Following the LLM-TLS method \cite{hu-etal-2024-moments}, we propose a new approach for the constrained timeline summarization task by leveraging a large language model (LLM) for summarization and clustering, which we call \textsc{REACTS} (\underline{RE}flective \underline{A}lgorithm for \underline{C}onstrained \underline{T}imeline \underline{S}ummarization). Our method consists of four main steps: event summarization, self-reflection, event clustering, and finally, cluster and sentence selection. The core idea is to summarize each document according to the constraint, cluster the summaries that relate to the same event, and transform the clusters into event descriptions with corresponding dates. We illustrate our method in Figure \ref{fig:method}.

\subsection{Event Summarization}
Inspired by the effectiveness of LLMs in query-based summarization \cite{jiang-etal-2024-trisum}, we employ large language models for event summarization. The summary is expected to be in the format of a date followed by a one-sentence summary of a key event in the article related to the keyword and that adheres to the constraint, such as, "2021-06-04: The miniseries *Lisey’s Story*, adapted by King and based on his 2006 novel of the same name, premieres on Apple TV+." If there is nothing to summarize that meets the constraint, the model is expected to output \texttt{NULL}. We refer to this process as \textsc{ConstrainedTopicSum} in Algorithm \ref{alg:method}.

Each article includes a publication date, but the important event may occur sometime before the publication date without an explicit mention of the exact date in the article. To assist the model in generating the correct date, we preprocess the news articles by prepending the sentence with the exact date whenever a time reference is mentioned. For example, if the publication date is 14 August 2024, and a sentence in the article contains a time reference like ``yesterday,'' the sentence is prepended with ``(2024-08-13)''. Similarly, if the article mentions ``last Friday,'' the sentence is prepended with ``(2024-08-09)''. The time references are parsed using HeidelTime\footnote{\url{https://github.com/HeidelTime/heideltime}} \cite{strotgen-gertz-2015-baseline}.

\subsection{Self-Reflection}
Self-evaluation techniques have been reported to improve the reasoning capabilities of LLMs \cite{weng-etal-2023-large, xie2023selfevaluation}. We observe that LLMs often produce an event summary even when it does not adhere to the specified constraint. To mitigate this, we employ self-reflection as an additional verification step by prompting the same LLM to assess whether the summary it just generated, $e_i$, for topic $t$ adheres to the constraint $c$. If the model determines that it does not, $e_i$ is discarded and excluded from the timeline generation. We refer to this process as \textsc{AdhereToConstraint} in Algorithm \ref{alg:method}. The input prompt to perform self-reflection is given in Table \ref{tab:prompt_self_reflect}.

\begin{table}[ht]
\centering
\begin{tabular}{|p{0.94\linewidth}|}
    \hline \\
    Review the timestamped event description related to {keyword}, accompanied by a constraint. Please determine whether the event description complies with or corresponds to the constraint. Respond with `Yes' if the event description aligns with the constraint, or with `No' if it does not.\\
\#\#\#\#\#\#\#\#\#\#\#\#\#\#\#\#\#\\
\{positive example\}\\
\\
\#\#\#\#\#\#\#\#\#\#\#\#\#\#\#\#\#\\
\{negative example\}\\
\\
\#\#\#\#\#\#\#\#\#\#\#\#\#\#\#\#\#\\
\#\#\# Event\\
\{event\}\\
\\
\#\#\# Constraint\\
\{constraint\}\\
\#\#\# Answer\\
    \hline
\end{tabular}
\caption{Prompt template for self-reflection.}
\label{tab:prompt_self_reflect}
\end{table}

\subsection{Event Clustering}
For every event summary $e_i$ that passes the self-reflection step, the event description is encoded using the General Text Embeddings (GTE) model \cite{li2023generaltextembeddingsmultistage}. The summaries are transformed into embedding vectors so that semantically similar event summaries describing the same event can be accurately grouped together into a cluster. GTE performs exceptionally well on the Massive Text Embedding Benchmark (MTEB) while maintaining a modest number of parameters, making it an ideal choice for encoding summaries from tens of thousands of articles.

\begin{table*}[t]
\centering
\begin{tabular}{l|l|rrr|rrr|rrr}
    \hline
    \hline
{} & {} & \multicolumn{3}{c|}{\textbf{AR-1}} & \multicolumn{3}{c|}{\textbf{AR-2}} & \multicolumn{3}{c}{\textbf{Date F1}}\\
\textbf{Model} & \textbf{LLM} & \textbf{P} & \textbf{R} & \textbf{F1} & \textbf{P} & \textbf{R} & \textbf{F1} & \textbf{P} & \textbf{R} & \textbf{F1} \\
    \hline
    \multicolumn{10}{l}{All events} \\
    \hline
    \hspace{1.5mm} \textsc{REACTS} w/o SR & L3.1-8B & 0.0580 & 0.0541 & 0.0561 & 0.0262 & 0.0264 & 0.0263 & 0.1399 & 0.1391 & 0.1395 \\
    \hspace{1.5mm} \textsc{REACTS} & L3.1-8B & \textbf{0.0859} & \textbf{0.0695} &  \textbf{0.0768} & \textbf{0.0381} & \textbf{0.0326} & \textbf{0.0351} & \textbf{0.1809} & \textbf{0.1710} & \textbf{0.1758} \\
    \hdashline
    \hspace{1.5mm} \textsc{REACTS} w/o SR & L3.1-70B & 0.0701 & 0.0957 & 0.0809 & 0.0326 & 0.0496 &  0.0393 & 0.1773 & 0.1773 & 0.1773  \\
    \hspace{1.5mm} \textsc{REACTS} & L3.1-70B & \textbf{0.0970} & \textbf{0.1246} & \textbf{0.1091} & \textbf{0.0434} & \textbf{0.0592} & \textbf{0.0501} & \textbf{0.2425} & \textbf{0.2396} & \textbf{0.2411}\\
    \hline
    \multicolumn{10}{l}{Filtered events} \\
    \hline
    \hspace{1.5mm} \textsc{Baseline} & L3.1-8B & 0.0387 & 0.0180 & 0.0246 & 0.0042 & 0.0026 & 0.0032 & 0.1005 & 0.0574 & 0.0731 \\
    \hspace{1.5mm} \textsc{REACTS} w/o SR & L3.1-8B & 0.0769 & 0.0693 & 0.0729 & 0.0337 & 0.0339 & 0.0338 &  0.1749 & 0.1743 & 0.1746 \\
    \hspace{1.5mm} \textsc{REACTS} & L3.1-8B & \textbf{0.1095} & \textbf{0.0882} & \textbf{0.0977} & \textbf{0.0497} & \textbf{0.0427} & \textbf{0.0459} & \textbf{0.2266} & \textbf{0.2211} & \textbf{0.2238} \\
    \hdashline
    \hspace{1.5mm} \textsc{Baseline} & L3.1-70B & 0.0687 & 0.0939 & 0.0793 & 0.0324 & 0.0480 & 0.0387 & 0.1341 & 0.2524 & 0.1751 \\
    \hspace{1.5mm} \textsc{REACTS} w/o SR & L3.1-70B & 0.0906 & 0.1220 & 0.1040 & 0.0405 & 0.0614 & 0.0488 & 0.2315 & 0.2315 & 0.2315 \\
    \hspace{1.5mm} \textsc{REACTS} & L3.1-70B & \textbf{0.1152} & \textbf{0.1533} & \textbf{0.1316} & \textbf{0.0483} & \textbf{0.0703} & \textbf{0.0572} & \textbf{0.2925} & \textbf{0.2925} & \textbf{0.2925} \\
    \hline
    \multicolumn{10}{l}{Filtered events (10\% data)} \\
    \hline
    \hspace{1.5mm} \textsc{Baseline} & GPT-4o & 0.0487 & \textbf{0.1451} & 0.0729 & 0.0216 & 0.0730 & 0.0334 & 0.2065 & \textbf{0.3176} & 0.2506 \\
    \hspace{1.5mm} \textsc{REACTS} & GPT-4o & \textbf{0.0652} & 0.1386 & \textbf{0.0887} & \textbf{0.0281} & \textbf{0.0752} & \textbf{0.0409} & \textbf{0.3000} & 0.3000 & \textbf{0.3000} \\
    \hline
    \hline
\end{tabular}
\caption{Score comparison of the baseline method, our method (\textsc{REACTS}), our method without self-reflection (\textsc{REACTS} w/o SR) using Llama 3.1 8B (L3.1-8B), Llama 3.1 70B (L3.1-70B), and GPT-4o on our dataset (\textsc{CREST}). We evaluate the models on precision (P), recall (R), and F1 scores using alignment-based ROUGE-1 (AR-1), alignment-based ROUGE-2 (AR-2), and date F1-score metrics. The best scores for each experiment setting are boldfaced.}
\label{tab:main_result}
\end{table*}

To generate clusters, from a vector database $D$, we retrieve $N$ event descriptions that have the closest embedding vectors to the current event description being processed (the \textsc{Retrieve} function in Algorithm \ref{alg:method}). For each pair consisting of the current event description and its retrieved neighbor, we use an LLM with few-shot prompting to check whether they describe the same event. In addition to this description matching by the LLM, we also check whether the event dates are the same, as events with similar descriptions but different dates likely represent distinct occurrences. This process is denoted as the \textsc{SameEvent} function in Algorithm \ref{alg:method}. If the pair passes both checks, the current event description is added to the cluster of its first matching neighbor. If the event description does not match any of its top $N$ neighbors, it forms its own cluster. Finally, the embedding vector of each summary is stored in $D$ to facilitate grouping with similar subsequent event descriptions.

\subsection{Cluster and Sentence Selection}
Each cluster represents an event related to the topic. To generate a timeline with $l$ events, we select the best $l$ clusters and summarize the event descriptions within each cluster into $k$ sentences as specified by the user. We employ a heuristic to choose the top $l$ clusters based on size, with the intuition that more significant events are covered by more articles, especially in the news domain. Subsequently, we apply TextRank \cite{mihalcea-tarau-2004-textrank} to select the best $k$ sentences within each cluster.

\subsection{Baseline Method}
A straightforward approach to perform constrained timeline summarization using an LLM is to concatenate the articles into the prompt and directly ask the model to produce a constrained timeline. However, LLMs have a limited context window size, so it is not always possible to fit the entire article pool within the input. To address this limitation, the baseline method involves randomly sampling articles and incrementally adding them to the input prompt one by one until the input context size limit is reached (taking into account the space needed for the instruction prompt and the output). Next, an instruction is added to the prompt, asking the model to generate a timeline comprising $l$ events, each described with a date and a $k$-sentence summary that adheres to the constraint $c$. The model then generates the timeline token-by-token until it either determines that it should stop or reaches the token limit.

\section{Experiments}
We run experiments to investigate whether self-reflection helps in generating more relevant timelines. We evaluate our method on our proposed dataset, both against the ground-truth timelines with all events and ground-truth timelines with filtered events. We employ Llama-3.1 8B\footnote{\url{https://llama.meta.com/}} \cite{dubey2024llama3herdmodels}, Llama-3.1 70B, and GPT-4o \cite{openai2024gpt4technicalreport} as the LLMs for our proposed method and the baseline method. However, we only evaluate models with GPT-4o on 10\% of the test set due to cost consideration. In all experiments, we set the generation temperature of the LLMs to zero to make the results reproducible.

As previously explained, the baseline method is inherently limited by the maximum context length of the LLM. Therefore, it can only consider a limited number of articles when generating a timeline. To evaluate the best possible performance of the baseline method, we also conduct additional experiments with an oracle article retriever. From the article pool, the oracle retriever retrieves only articles relevant to the events in the unconstrained ground-truth timeline. Even though the oracle retriever helps to filter out noisy articles, the models still need to determine whether the events in the articles adhere to the constraints. The set of articles kept by the oracle retriever is then randomly sampled to fit the context length of the baseline method and used as the final article pool for all methods. That is, in this experiment, all methods receive the exact same set of input articles to generate the timeline summary. We use GPT-4o as the oracle retriever; therefore, we do not use it as the backbone LLM for the methods.

\subsection{Evaluation}
We evaluate the experiments with the standard metrics for the timeline summarization task, which are alignment-based ROUGE F1-score \cite{martschat-markert-2017-improving} and date F1-score \cite{martschat-markert-2018-temporally}. We employ an approximate randomization test \cite{riezler-maxwell-2005-pitfalls, chinchor-etal-1993-evaluating} with 100 trials and a $p$-value of 0.05 to measure statistical significance.

\begin{table*}[t]
\centering
\begin{tabular}{l|l|rrr|rrr|rrr}
    \hline
    \hline
{} & {} & \multicolumn{3}{c|}{\textbf{AR-1}} & \multicolumn{3}{c|}{\textbf{AR-2}} & \multicolumn{3}{c}{\textbf{Date F1}}\\
\textbf{Model} & \textbf{LLM} & \textbf{P} & \textbf{R} & \textbf{F1} & \textbf{P} & \textbf{R} & \textbf{F1} & \textbf{P} & \textbf{R} & \textbf{F1} \\
    \hline
    \multicolumn{10}{l}{With oracle retriever on filtered events} \\
    \hline
    \hspace{1.5mm} \textsc{Baseline} & L3.1-8B & {0.0761} & {0.0829} & 0.0794 & {0.0299} & {0.0310} & 0.0305 & {0.2160} & \textbf{0.2935} & 0.2489 \\
    \hspace{1.5mm} \textsc{REACTS} w/o SR & L3.1-8B & {0.0966} & {0.0800} & 0.0875 & {0.0399} & {0.0341} & 0.0367 & {0.2172} & {0.2151} & 0.2161 \\
    \hspace{1.5mm} \textsc{REACTS} & L3.1-8B & \textbf{0.1505} & \textbf{0.1102} & \textbf{0.1272} & \textbf{0.0687} & \textbf{0.0490} & \textbf{0.0572} & \textbf{0.2916} & {0.2766} & \textbf{0.2839} \\
    \hdashline
    \hspace{1.5mm} \textsc{Baseline} & L3.1-70B & {0.1149} & {0.1764} & 0.1392 & {0.0429} & {0.0767} & 0.0550 & {0.2539} & \textbf{0.4811} & 0.3324 \\
    \hspace{1.5mm} \textsc{REACTS} w/o SR & L3.1-70B & {0.1142} & {0.1488} & 0.1292 & {0.0482} & {0.0655} & 0.0555 & {0.3165} & {0.3132} & 0.3148 \\
    \hspace{1.5mm} \textsc{REACTS} & L3.1-70B & \textbf{0.1810} & \textbf{0.2213} & \textbf{0.1991} & \textbf{0.0735} & \textbf{0.0959} & \textbf{0.0832} & \textbf{0.4769} & {0.4485} & \textbf{0.4622} \\
    \hline
    \hline
\end{tabular}
\caption{Results of the experiments with oracle retriever of the baseline method, our method (\textsc{REACTS}), our method without self-reflection (\textsc{REACTS} w/o SR) using Llama 3.1 8B (L3.1-8B) and Llama 3.1 70B (L3.1-70B). We evaluate the models on precision (P), recall (R), and F1 scores using alignment-based ROUGE-1 (AR-1), alignment-based ROUGE-2 (AR-2), and date F1-score metrics. The best scores for each experiment setting are boldfaced.}
\label{tab:oracle_result}
\end{table*}

\subsubsection{Alignment-Based ROUGE F1-score}
The alignment-based ROUGE F1-score measures the text overlap of the event descriptions between the predicted timeline and the ground-truth timeline. It first aligns the events in the predicted timeline with events in the ground-truth timeline based on the closeness of the dates and the similarity of the event descriptions. Following \cite{gholipour-ghalandari-ifrim-2020-examining}, we use the alignment setting that allows many-to-one alignment.

For each pair of aligned predicted event and ground-truth event, the metric\footnote{We use ROUGE v1.5.5 by Chin-Yew Lin with \texttt{-n 2 -m -s} arguments, which measures up to 2-gram similarity of stemmed words, ignoring stopwords.} measures the n-gram similarity between the event descriptions. Precision is proportional to the ratio of the overlap compared to the predicted event description, while recall is proportional to the ratio of the overlap compared to the ground-truth event description.

\subsubsection{Date F1-Score}
The date F1-score simply measures the F1 score of the dates covered in the ground-truth timeline against the dates in the predicted timeline. Unlike alignment-based ROUGE, date F1-score performs hard matching of the dates and does not consider the event descriptions.

\section{Results}
We present our main experimental results in Table \ref{tab:main_result}. Our findings indicate that our method significantly outperforms the baseline. The baseline method using Llama struggles to produce a coherent timeline. It often fails to determine when to stop and occasionally generates nonsensical outputs, especially with Llama-3.1 8B. Even when we use GPT-4o, our method still achieves better F1 scores than the baseline. However, note that GPT-4o may have a slight advantage over the other LLMs, as it was used in the dataset creation process.

We also observe that self-reflection significantly improves all scores (i.e., precision, recall, and F1) across all metrics (i.e., AR-1, AR-2, and date F1) in all experimental settings with Llama-3.1. With Llama-3.1 70B, when evaluated against ground-truth timelines without event filtering (all events), self-reflection improves the AR-1 F1 score by 2.82\%, the AR-2 F1 score by 1.08\%, and the date F1 score by 6.38\%. When evaluated against filtered ground-truth timelines, self-reflection improves the AR-1 F1 score by 2.76\%, the AR-2 F1 score by 0.84\%, and the date F1 score by 6.10\%.

With the oracle retriever (Table \ref{tab:oracle_result}), using Llama-3.1 8B, our method still significantly outperforms the baseline by 4.78\%, 2.67\%, and 3.50\% on AR-1 F1, AR-2 F1, and date F1 respectively. The score improvements are even greater with the larger Llama-3.1 70B model, reaching 5.99\%, 2.82\%, and 12.98\% on AR-1 F1, AR-2 F1, and date F1 respectively.

The baseline method is impractical for real-world applications where hundreds of thousands of news articles are published each month. Regardless of the context window size, it cannot keep up with the speed and volume of information flowing through the internet. We have shown that even with the same set of articles (in the oracle retriever setup), our method is superior. Furthermore, the baseline method is unsuitable for online (streaming) processing. Every time a new article is added, previous articles need to be reprocessed to update the timeline, leading to significant computational inefficiency.

\section{Conclusion}
In this paper, we propose a new task with high relevance to current needs, called constrained timeline summarization. We present a new test set for the task (\textsc{CREST}), which was built by generating the constraints using human-in-the-loop collaboration with an LLM, hiring annotators to annotate the adherence of the events in the ground-truth timeline to the constraints, and filtering the events without supporting articles by utilizing an LLM.

We also propose an effective method that utilizes LLMs for the task. Our method does not require any training and can work with different LLMs. Our method works by summarizing the articles according to the constraint, employing a self-reflection procedure to filter out irrelevant summaries, clustering the summaries that describe the same event, and selecting the top $l$ clusters and the top $k$ sentences for each cluster to generate the constrained timeline.

Lastly, we demonstrate the effectiveness of our method by comparing it against a baseline method that generates the timeline directly by concatenating all the articles into its input prompt. We show that our method successfully outperforms the baseline on all metrics. Similarly, we demonstrate the effectiveness of self-reflection by comparing our method to a variant of our method that does not employ self-reflection, and show that self-reflection effectively improves the F1 scores on all metrics. With this work, we hope that constrained timeline summarization can gain more attention and more progress can be achieved on this task in future.

\section{Acknowledgments}

This work is fully supported by the Advanced Research and Technology Innovation Centre (ARTIC), the National University of Singapore under Grant (project number: ELDT-RP1).

\bibliography{aaai25}

\begin{thebibliography}{40}
\providecommand{\natexlab}[1]{#1}

\bibitem[{Allan, Gupta, and Khandelwal(2001)}]{10.1145/383952.383954}
Allan, J.; Gupta, R.; and Khandelwal, V. 2001.
\newblock Temporal summaries of new topics.
\newblock In \emph{Proceedings of SIGIR}, 10--18.

\bibitem[{Cao et~al.(2024)Cao, Jiao, Yan, Zhang, Tang, and Zhuang}]{cao2024idealleveraginginfinitedynamic}
Cao, J.; Jiao, D.; Yan, Q.; Zhang, W.; Tang, S.; and Zhuang, Y. 2024.
\newblock {IDEAL}: Leveraging infinite and dynamic characterizations of large language models for query-focused summarization.
\newblock arXiv:2407.10486.

\bibitem[{Chieu and Lee(2004)}]{10.1145/1008992.1009065}
Chieu, H.~L.; and Lee, Y.~K. 2004.
\newblock Query based event extraction along a timeline.
\newblock In \emph{Proceedings of SIGIR}, 425--432.

\bibitem[{Chinchor, Hirschman, and Lewis(1993)}]{chinchor-etal-1993-evaluating}
Chinchor, N.; Hirschman, L.; and Lewis, D.~D. 1993.
\newblock Evaluating message understanding systems: An analysis of the third {M}essage {U}nderstanding {C}onference ({MUC}-3).
\newblock \emph{Computational Linguistics}, 19(3): 409--450.

\bibitem[{Conroy, Schlesinger, and O{'}Leary(2006)}]{conroy-etal-2006-topic}
Conroy, J.~M.; Schlesinger, J.~D.; and O{'}Leary, D.~P. 2006.
\newblock Topic-focused multi-document summarization using an approximate oracle score.
\newblock In \emph{Proceedings of {COLING}/{ACL}}, 152--159.

\bibitem[{Dang and Owczarzak(2009)}]{tac08}
Dang, H.; and Owczarzak, K. 2009.
\newblock Overview of the {TAC} 2008 update summarization task.
\newblock In \emph{Proceedings of TAC}.

\bibitem[{Duan, Jatowt, and Yoshikawa(2020)}]{Duan2020ComparativeTS}
Duan, Y.; Jatowt, A.; and Yoshikawa, M. 2020.
\newblock Comparative timeline summarization via dynamic affinity-preserving random walk.
\newblock In \emph{Proceedings of ECAI}, 1778--1785.

\bibitem[{Ghalandari and Ifrim(2020)}]{gholipour-ghalandari-ifrim-2020-examining}
Ghalandari, D.~G.; and Ifrim, G. 2020.
\newblock Examining the state-of-the-art in news timeline summarization.
\newblock In \emph{Proceedings of ACL}, 1322--1334.

\bibitem[{Gilardi, Alizadeh, and Kubli(2023)}]{Gilardi2023ChatGPTOC}
Gilardi, F.; Alizadeh, M.; and Kubli, M. 2023.
\newblock ChatGPT outperforms crowd workers for text-annotation tasks.
\newblock \emph{Proceedings of the National Academy of Sciences of the United States of America}, 120(30).

\bibitem[{Hamborg, Meuschke, and Gipp(2018)}]{Hamborg2018BiasawareNA}
Hamborg, F.; Meuschke, N.; and Gipp, B. 2018.
\newblock Bias-aware news analysis using matrix-based news aggregation.
\newblock \emph{International Journal on Digital Libraries}, 21: 129--147.

\bibitem[{Hu, Moon, and Ng(2024)}]{hu-etal-2024-moments}
Hu, Q.; Moon, G.; and Ng, H.~T. 2024.
\newblock From moments to milestones: Incremental timeline summarization leveraging large language models.
\newblock In \emph{Proceedings of ACL}, 7232--7246.

\bibitem[{Jiang et~al.(2024)Jiang, Xiao, Wang, Bhatia, Sun, and Han}]{jiang-etal-2024-trisum}
Jiang, P.; Xiao, C.; Wang, Z.; Bhatia, P.; Sun, J.; and Han, J. 2024.
\newblock {TriSum}: Learning summarization ability from large language models with structured rationale.
\newblock In \emph{Proceedings of NAACL}, 2805--2819.

\bibitem[{Krishna, Kumar, and Reddy(2013)}]{10.5120/11587-6925}
Krishna, R. V. V.~M.; Kumar, S. Y.~P.; and Reddy, C.~S. 2013.
\newblock A hybrid method for query based automatic summarization system.
\newblock \emph{International Journal of Computer Applications}, 68(6): 39--43.

\bibitem[{La~Quatra et~al.(2021)La~Quatra, Cagliero, Baralis, Messina, and Montagnuolo}]{10.1145/3404835.3462954}
La~Quatra, M.; Cagliero, L.; Baralis, E.; Messina, A.; and Montagnuolo, M. 2021.
\newblock Summarize dates first: A paradigm shift in timeline summarization.
\newblock In \emph{Proceedings of SIGIR}, 418--427.

\bibitem[{Laskar, Hoque, and Huang(2020)}]{Laskar2020QueryFA}
Laskar, M. T.~R.; Hoque, E.; and Huang, X. 2020.
\newblock Query focused abstractive summarization via incorporating query relevance and transfer learning with transformer models.
\newblock In \emph{Proceedings of Canadian AI}, 342--348.

\bibitem[{Lewis et~al.(2020)Lewis, Liu, Goyal, Ghazvininejad, Mohamed, Levy, Stoyanov, and Zettlemoyer}]{lewis-etal-2020-bart}
Lewis, M.; Liu, Y.; Goyal, N.; Ghazvininejad, M.; Mohamed, A.; Levy, O.; Stoyanov, V.; and Zettlemoyer, L. 2020.
\newblock {BART}: Denoising sequence-to-sequence pre-training for natural language generation, translation, and comprehension.
\newblock In \emph{Proceedings of ACL}, 7871--7880.

\bibitem[{Li et~al.(2021)Li, Ma, Yu, Wu, Gao, Ji, and McKeown}]{li-etal-2021-timeline}
Li, M.; Ma, T.; Yu, M.; Wu, L.; Gao, T.; Ji, H.; and McKeown, K. 2021.
\newblock Timeline summarization based on event graph compression via time-aware optimal transport.
\newblock In \emph{Proceedings of EMNLP}, 6443--6456.

\bibitem[{Li et~al.(2023)Li, Zhang, Zhang, Long, Xie, and Zhang}]{li2023generaltextembeddingsmultistage}
Li, Z.; Zhang, X.; Zhang, Y.; Long, D.; Xie, P.; and Zhang, M. 2023.
\newblock Towards general text embeddings with multi-stage contrastive learning.
\newblock arXiv:2308.03281.

\bibitem[{Liu and Lapata(2019)}]{liu-lapata-2019-text}
Liu, Y.; and Lapata, M. 2019.
\newblock Text summarization with pretrained encoders.
\newblock In \emph{Proceedings of EMNLP}, 3730--3740.

\bibitem[{Llama~Team(2024)}]{dubey2024llama3herdmodels}
Llama~Team, A. 2024.
\newblock The {L}lama 3 herd of models.
\newblock arXiv:2407.21783.

\bibitem[{Mani and Bloedorn(1998)}]{Mani1998MachineLO}
Mani, I.; and Bloedorn, E. 1998.
\newblock Machine learning of generic and user-focused summarization.
\newblock In \emph{Proceedings of AAAI}, 821--826.

\bibitem[{Martschat and Markert(2017)}]{martschat-markert-2017-improving}
Martschat, S.; and Markert, K. 2017.
\newblock Improving {ROUGE} for timeline summarization.
\newblock In \emph{Proceedings of EACL}, 285--290.

\bibitem[{Martschat and Markert(2018)}]{martschat-markert-2018-temporally}
Martschat, S.; and Markert, K. 2018.
\newblock A temporally sensitive submodularity framework for timeline summarization.
\newblock In \emph{Proceedings of {CoNLL}}, 230--240.

\bibitem[{Mihalcea and Tarau(2004)}]{mihalcea-tarau-2004-textrank}
Mihalcea, R.; and Tarau, P. 2004.
\newblock {TextRank}: Bringing order into text.
\newblock In \emph{Proceedings of EMNLP}, 404--411.

\bibitem[{Mohamed and Rajasekaran(2006)}]{4042277}
Mohamed, A.~A.; and Rajasekaran, S. 2006.
\newblock Improving query-based summarization using document graphs.
\newblock In \emph{Proceedings of ISSPIT}, 408--410.

\bibitem[{OpenAI(2024)}]{openai2024gpt4technicalreport}
OpenAI. 2024.
\newblock {GPT}-4 technical report.
\newblock arXiv:2303.08774.

\bibitem[{Ouyang et~al.(2011)Ouyang, Li, Li, and Lu}]{OUYANG2011227}
Ouyang, Y.; Li, W.; Li, S.; and Lu, Q. 2011.
\newblock Applying regression models to query-focused multi-document summarization.
\newblock \emph{Information Processing and Management}, 47(2): 227--237.

\bibitem[{Park and Ko(2022)}]{10.1145/3477495.3531901}
Park, C.; and Ko, Y. 2022.
\newblock {QSG T}ransformer: Transformer with query-attentive semantic graph for query-focused summarization.
\newblock In \emph{Proceedings of SIGIR}, 2589--2594.

\bibitem[{Rahman and Borah(2015)}]{7377323}
Rahman, N.; and Borah, B. 2015.
\newblock A survey on existing extractive techniques for query-based text summarization.
\newblock In \emph{Proceedings of ISACC}, 98--102.

\bibitem[{Riezler and Maxwell(2005)}]{riezler-maxwell-2005-pitfalls}
Riezler, S.; and Maxwell, J.~T. 2005.
\newblock On some pitfalls in automatic evaluation and significance testing for {MT}.
\newblock In \emph{Proceedings of the {ACL} Workshop on Intrinsic and Extrinsic Evaluation Measures for Machine Translation and/or Summarization}, 57--64.

\bibitem[{Steinberger and Je\v{z}ek(2009)}]{10.1145/1600193.1600239}
Steinberger, J.; and Je\v{z}ek, K. 2009.
\newblock Update summarization based on novel topic distribution.
\newblock In \emph{Proceedings of DocEng}, 205--213.

\bibitem[{Str{\"o}tgen and Gertz(2015)}]{strotgen-gertz-2015-baseline}
Str{\"o}tgen, J.; and Gertz, M. 2015.
\newblock A baseline temporal tagger for all languages.
\newblock In \emph{Proceedings of EMNLP}, 541--547.

\bibitem[{Touvron et~al.(2023)Touvron, Martin, Stone, Albert, Almahairi, Babaei, Bashlykov, Batra, Bhargava, Bhosale, Bikel, Blecher, Ferrer, Chen, Cucurull, Esiobu, Fernandes, Fu, Fu, Fuller, Gao, Goswami, Goyal, Hartshorn, Hosseini, Hou, Inan, Kardas, Kerkez, Khabsa, Kloumann, Korenev, Koura, Lachaux, Lavril, Lee, Liskovich, Lu, Mao, Martinet, Mihaylov, Mishra, Molybog, Nie, Poulton, Reizenstein, Rungta, Saladi, Schelten, Silva, Smith, Subramanian, Tan, Tang, Taylor, Williams, Kuan, Xu, Yan, Zarov, Zhang, Fan, Kambadur, Narang, Rodriguez, Stojnic, Edunov, and Scialom}]{touvron2023llama2openfoundation}
Touvron, H.; Martin, L.; Stone, K.; Albert, P.; Almahairi, A.; Babaei, Y.; Bashlykov, N.; Batra, S.; Bhargava, P.; Bhosale, S.; Bikel, D.; Blecher, L.; Ferrer, C.~C.; Chen, M.; Cucurull, G.; Esiobu, D.; Fernandes, J.; Fu, J.; Fu, W.; Fuller, B.; Gao, C.; Goswami, V.; Goyal, N.; Hartshorn, A.; Hosseini, S.; Hou, R.; Inan, H.; Kardas, M.; Kerkez, V.; Khabsa, M.; Kloumann, I.; Korenev, A.; Koura, P.~S.; Lachaux, M.-A.; Lavril, T.; Lee, J.; Liskovich, D.; Lu, Y.; Mao, Y.; Martinet, X.; Mihaylov, T.; Mishra, P.; Molybog, I.; Nie, Y.; Poulton, A.; Reizenstein, J.; Rungta, R.; Saladi, K.; Schelten, A.; Silva, R.; Smith, E.~M.; Subramanian, R.; Tan, X.~E.; Tang, B.; Taylor, R.; Williams, A.; Kuan, J.~X.; Xu, P.; Yan, Z.; Zarov, I.; Zhang, Y.; Fan, A.; Kambadur, M.; Narang, S.; Rodriguez, A.; Stojnic, R.; Edunov, S.; and Scialom, T. 2023.
\newblock {L}lama 2: {O}pen foundation and fine-tuned chat models.
\newblock arXiv:2307.09288.

\bibitem[{Tran, Herder, and Markert(2015)}]{tran-etal-2015-joint}
Tran, G.; Herder, E.; and Markert, K. 2015.
\newblock Joint graphical models for date selection in timeline summarization.
\newblock In \emph{Proceedings of ACL}, 1598--1607.

\bibitem[{Wang et~al.(2013)Wang, Li, Li, Li, and Wei}]{WANG2013271}
Wang, W.; Li, S.; Li, J.; Li, W.; and Wei, F. 2013.
\newblock Exploring hypergraph-based semi-supervised ranking for query-oriented summarization.
\newblock \emph{Information Sciences}, 237: 271--286.

\bibitem[{Weng et~al.(2023)Weng, Zhu, Xia, Li, He, Liu, Sun, Liu, and Zhao}]{weng-etal-2023-large}
Weng, Y.; Zhu, M.; Xia, F.; Li, B.; He, S.; Liu, S.; Sun, B.; Liu, K.; and Zhao, J. 2023.
\newblock Large language models are better reasoners with self-verification.
\newblock In \emph{Findings of EMNLP}, 2550--2575.

\bibitem[{Xie et~al.(2024)Xie, Kawaguchi, Zhao, Zhao, Kan, He, and Xie}]{xie2023selfevaluation}
Xie, Y.; Kawaguchi, K.; Zhao, Y.; Zhao, J.~X.; Kan, M.-Y.; He, J.; and Xie, M.~Q. 2024.
\newblock Self-evaluation guided beam search for reasoning.
\newblock In \emph{Proceedings of NeurIPS}, 41618--41650.

\bibitem[{Xu et~al.(2023)Xu, Wang, Liu, Wang, Xu, Iter, He, Zhu, and Zeng}]{xu-etal-2023-lmgqs}
Xu, R.; Wang, S.; Liu, Y.; Wang, S.; Xu, Y.; Iter, D.; He, P.; Zhu, C.; and Zeng, M. 2023.
\newblock {LMGQS}: A large-scale dataset for query-focused summarization.
\newblock In \emph{Findings of EMNLP}, 14764--14776.

\bibitem[{You et~al.(2022)You, Li, Kamigaito, Funakoshi, and Okumura}]{you-etal-2022-joint}
You, J.; Li, D.; Kamigaito, H.; Funakoshi, K.; and Okumura, M. 2022.
\newblock Joint learning-based heterogeneous graph attention network for timeline summarization.
\newblock In \emph{Proceedings of NAACL}, 4091--4104.

\bibitem[{Yu et~al.(2021)Yu, Jatowt, Doucet, Sugiyama, and Yoshikawa}]{yu-etal-2021-multi}
Yu, Y.; Jatowt, A.; Doucet, A.; Sugiyama, K.; and Yoshikawa, M. 2021.
\newblock Multi-{T}ime{L}ine Summarization ({MTLS}): Improving timeline summarization by generating multiple summaries.
\newblock In \emph{Proceedings of ACL}, 377--387.

\end{thebibliography}

\clearpage

\appendix
\section{Appendix}
\section{Annotation Instruction}
\begin{figure*}[t]
\centering
\includegraphics[width=0.95\textwidth]{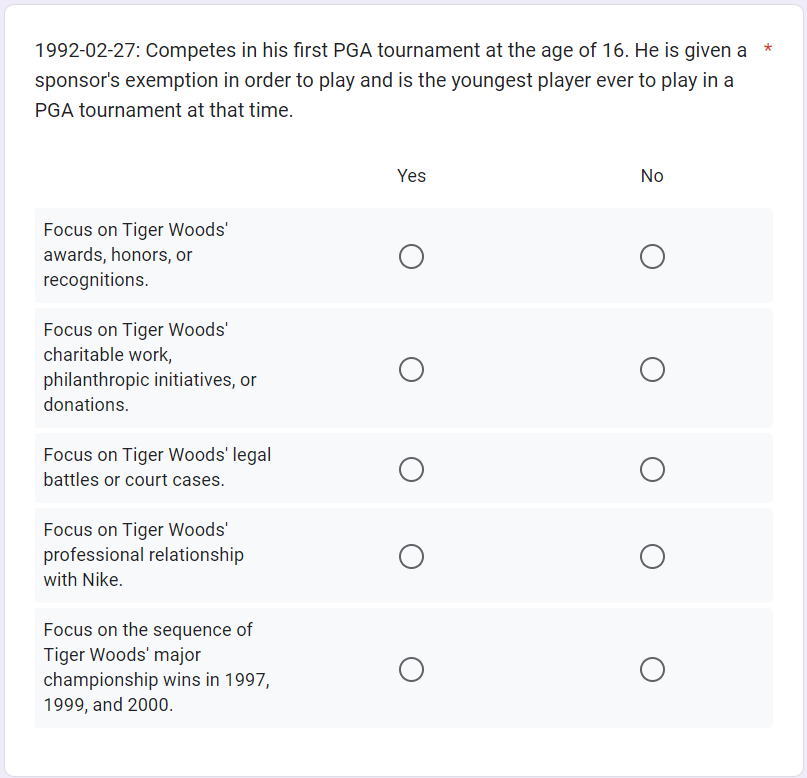}
\caption{Example of the annotation interface for the event-constraint pairs in Tiger Woods's timeline.}
\label{fig:annot_interface}
\end{figure*}

Before performing the event-constraint annotation, all annotators are required to read the annotation guidelines provided in Table \ref{tab:annot_instruct}. They will then perform the annotation by answering ``yes'' or ``no'' for each event-constraint pair, indicating whether the event adheres to the constraint. The annotation interface is shown in Figure \ref{fig:annot_interface}.

\begin{table}[ht]
\centering
\begin{tabular}{p{0.94\linewidth}}
    \hline \\
\textbf{Instructions} \\
The goal of this annotation is to identify events in a timeline that satisfy certain constraints.\\
1. For an event in the timeline and each constraint (shown on Google Form), please click the 'Yes' button if the event meets the constraint, and 'No' if it does not. \\
2. The events in the timeline are real events reported by journalists about public figures or institutions. \\
3. We try to make the constraints as clear as possible, but sometimes they may require common knowledge (e.g., Obama was the president of the United States or California is
in the United States). Please annotate the events based on your best knowledge, without making any unnecessary assumptions. \\
4. Annotating the events should not require internet searches, but you are free to do so if needed. \\
5. An event may satisfy more than one constraint (Example 2) or none at all (Example 3). \\
6. Please refer to the examples below to see how some sample events are annotated in relation to each of five constraints. \\
    \hline \\
\end{tabular}
\caption{Annotation instruction.}
\label{tab:annot_instruct}
\end{table}

\section{Experimental Details}

\subsection{Implementation Details}
We use the vllm library\footnote{\url{https://github.com/vllm-project/vllm}} (version 0.5.4) to serve the local language models (Llama 3.1 8B and Llama 3.1 70B) and OpenAI Python API\footnote{\url{https://github.com/openai/openai-python}} library (version 0.28.1) to communicate with the language models. The GPT-4o model used in our experiments is \texttt{gpt-4o-2024-08-06}.

\subsection{Compute Resource}
We deploy Llama 3.1 8B on 2x NVIDIA A100 80GB GPUs (server \#1 in Table \ref{tab:servers}) and Llama 3.1 70B on 4x NVIDIA H100 80GB GPUs (server \#2 in Table \ref{tab:servers}). All experimental code was run on a server with 2x AMD EPYC 9554P CPUs and 792 GB of memory (server \#1).

\subsection{Prompts}
We provide the prompt templates for our experiments in the tables below.
\subsubsection{Summary Generation} Table \ref{tab:prompt_summary}
\subsubsection{Self-Reflection} Table \ref{tab:prompt_self_reflect_app}
\subsubsection{Event Similarity} Table \ref{tab:prompt_similarity}
\subsubsection{Baseline method} Table \ref{tab:prompt_baseline}

\subsection{Hyper-Parameters}
Our method does not require any training or fine-tuning, so we only need to set the decoding hyper-parameters for the language model. For clustering, the only hyper-parameter is determining the number of nearby events to retrieve from the vector database. We used the hyper-parameters of LLM-TLS \cite{hu-etal-2024-moments} and did not perform a hyper-parameter search ourselves.

\begin{table}[bth]
\centering
\begin{tabular}{l|lll}
    \hline
    Hyper-parameter & Value \\
    \hline
    Temperature & 0. \\
    Top-p & 1. \\
    \underline{Max. tokens} & {} \\
    - Summary generation & 256 \\
    - Self-reflect & 256 \\
    - Event similarity & 2 \\
    $N$ & 20 \\
    \hline    
\end{tabular}
\caption{The final hyper-parameters of our method. ``Max. tokens'' refers to the maximum number of tokens generated by the large language model, and $N$ refers to the number of neighboring events retrieved during clustering.}
\label{tab:hyper-param}
\end{table}

\begin{table*}[bth]
\centering
\begin{tabular}{l|lll}
    \hline
    Server & GPU & CPU & RAM \\
    \hline
    \#1 & 4x NVIDIA A100 80GB & 2x AMD EPYC 9554P & 792 GB \\
    \#2 & 4x NVIDIA H100 80GB & 2x INTEL XEON 8562Y+ & 528 GB \\
    \hline    
\end{tabular}
\caption{Specifications of the servers used in our experiments. Server \#1 has 4x NVIDIA A100 80GB GPUs, but we used only two GPUs throughout our experiments.}
\label{tab:servers}
\end{table*}

\begin{table*}[t]
\centering
\begin{tabular}{l|l|rrr}
    \hline
    \hline
\textbf{Model} & \textbf{LLM} & {\textbf{AR-1}} & {\textbf{AR-2}} & {\textbf{Date F1}}\\
    \hline
    \multicolumn{5}{l}{All events} \\
    \hline
    \hspace{1.5mm} \textsc{REACTS} w/o SR & L3.1-8B & $0.0575 \pm 0.0012$ & $0.0271 \pm 0.0007$ & $0.1415 \pm 0.0018$ \\
    \hspace{1.5mm} \textsc{REACTS} & L3.1-8B & $0.0770 \pm 0.0004$ & $0.0351 \pm 0.0004$ & $0.1761 \pm 0.0004$ \\
    \hdashline
    \hspace{1.5mm} \textsc{REACTS} w/o SR & L3.1-70B & $0.0804 \pm 0.0005$ & $0.0388 \pm 0.0005$ & $0.1771 \pm 0.0004$  \\
    \hspace{1.5mm} \textsc{REACTS} & L3.1-70B & $\textbf{0.1074} \pm 0.0015$ & $\textbf{0.0488} \pm 0.0011$ & $\textbf{0.2407} \pm 0.0006$\\
    \hline
    \multicolumn{5}{l}{Filtered events} \\
    \hline
    \hspace{1.5mm} \textsc{Baseline} & L3.1-8B & $0.0248 \pm 0.0008$ & $0.0047 \pm 0.0013$ & $0.0857 \pm 0.0110$ \\
    \hspace{1.5mm} \textsc{REACTS} w/o SR & L3.1-8B & $0.0738 \pm 0.0007$ & $0.0339 \pm 0.0002$ &  $0.1767 \pm 0.0019$ \\
    \hspace{1.5mm} \textsc{REACTS} & L3.1-8B & $0.0976 \pm 0.0007$ & $0.0456 \pm 0.0006$ & $0.2233 \pm 0.0010$ \\
    \hdashline
    \hspace{1.5mm} \textsc{Baseline} & L3.1-70B & $0.0871 \pm 0.0068$ & $0.0407 \pm 0.0018$ & $0.1866 \pm 0.0102$ \\
    \hspace{1.5mm} \textsc{REACTS} w/o SR & L3.1-70B & $0.1035 \pm 0.0006$ & $0.0484 \pm 0.0004$ & $0.2307 \pm 0.0014$ \\
    \hspace{1.5mm} \textsc{REACTS} & L3.1-70B & $\textbf{0.1311} \pm 0.0006$ & $\textbf{0.0566} \pm 0.0005$ & $\textbf{0.2923} \pm 0.0013$ \\
    \hline
    \hline
\end{tabular}
\caption{Average and standard deviation ($\bar{x} \pm \sigma$) of the F1 scores from our main experimental results.}
\label{tab:all_results}
\end{table*}

\section{Additional Results}
To ensure reproducibility, we ran the large language model with the decoding temperature set to 0. We conducted our experiments over three trials using three different few-shot examples for clustering, which are provided as JSON files in our code and can be loaded with the \texttt{-{}-few\_shot} argument when running the code. For the baseline method, we set the random seed for sampling the articles by specifying the \texttt{-{}-seed} argument in the code. We report the average and standard deviation in Table \ref{tab:all_results}.

We evaluated the statistical significance of our results using approximate randomization \cite{riezler-maxwell-2005-pitfalls, chinchor-etal-1993-evaluating} with 100 trials. Notably, we found that \textsc{REACTS} (with self-reflection) demonstrated statistically significant improvements ($p < 0.05$) compared to \textsc{REACTS} without self-reflection and the \textsc{Baseline} method across all experimental settings.

\begin{table}[ht]
\centering
\begin{tabular}{p{0.94\linewidth}}
    \hline \\
\#\#\# Instruction\\
Review the news article associated with the provided keyword and constraint. If the article's content does not relate to the keyword and specified constraint, output 'None'. Otherwise, summarize the most significant event related to the keyword while adhering to the constraint.\\
\\
\#\#\# Format\\
YYYY-MM-DD: One-sentence Summary\\
\\
\#\#\#\#\#\#\#\#\#\#\#\#\#\#\#\#\#\\
\#\#\# Keyword\\
Stephen King\\
\\
\#\#\# Constraint\\
Focus on Stephen King's book releases. \\
\\
\#\#\# Content\\
\textit{(example article about the release of the TV series adaptation of ``Lisey's story'', based on Stephen King's book)}\\
\\
\#\#\# Related Event Summary\\
None.\\
\\
\#\#\#\#\#\#\#\#\#\#\#\#\#\#\#\#\#\\
\#\#\# Keyword\\
Stephen King\\
\\
\#\#\# Constraint\\
Focus on Stephen King's involvement in television and streaming projects.\\
\\
\#\#\# Content\\
\textit{(example article about the release of the TV series adaptation of ``Lisey's story'', based on Stephen King's book)} \\
\\
\#\#\# Related Event Summary\\
2021-06-04: The miniseries “Lisey’s Story,” adapted by King and based on his 2006 novel of the same name, premieres on Apple TV+. \\
\\
\#\#\#\#\#\#\#\#\#\#\#\#\#\#\#\#\#\\
\#\#\# Keyword\\
\{topic keyword\}\\
\\
\#\#\# Constraint\\
\{constraint\}\\
\\
\#\#\# Content\\
\{content\}\\
\\
\#\#\# Related Event Summary \\
{} \\
    \hline \\
\end{tabular}
\caption{Prompt template for summary generation.}
\label{tab:prompt_summary}
\end{table}

\begin{table}[ht]
\centering
\begin{tabular}{p{0.94\linewidth}}
    \hline \\
    Review the timestamped event description related to {keyword}, accompanied by a constraint. Please determine whether the event description complies with or corresponds to the constraint. Respond with `Yes' if the event description aligns with the constraint, or with `No' if it does not.\\
\#\#\#\#\#\#\#\#\#\#\#\#\#\#\#\#\#\\
\{positive example\}\\
\\
\#\#\#\#\#\#\#\#\#\#\#\#\#\#\#\#\#\\
\{negative example\}\\
\\
\#\#\#\#\#\#\#\#\#\#\#\#\#\#\#\#\#\\
\#\#\# Event\\
\{event\}\\
\\
\#\#\# Constraint\\
\{constraint\}\\
\#\#\# Answer\\
{} \\
    \hline \\
\end{tabular}
\caption{Prompt template for self-reflection.}
\label{tab:prompt_self_reflect_app}
\end{table}

\begin{table}[ht]
\centering
\begin{tabular}{p{0.94\linewidth}}
    \hline \\
Taking the timestamps into account, evaluate whether two prior news events are referring to the same event related to the keyword. If the two events occur on the same date, and they are about the same topic related to the keyword, then they should be considered as referring to the same event. If so, please respond directly with `yes'. If not, respond with `no'. Then explain your answer. \\
-{}-{}-{}- \\
\{example \#1\} \\
-{}-{}-{}- \\
\{example \#2\} \\
-{}-{}-{}- \\
\{example \#3\} \\
-{}-{}-{}- \\
\# Keyword \\
\{topic keyword\} \\
\# Event 1 \\
\{event 1\} \\
\# Event 2 \\
\{event2\} \\
\# Answer \\
{} \\
    \hline \\
\end{tabular}
\caption{Prompt template for event similarity checking.}
\label{tab:prompt_similarity}
\end{table}

\begin{table}[ht]
\centering
\begin{tabular}{p{0.94\linewidth}}
    \hline \\
    \{article \#1\} \\
    \#\#\#\#\#\#\#\#\#\#\#\#\#\#\#\#\# \\
    ... \\
    \{article \#n\} \\
    \#\#\#\#\#\#\#\#\#\#\#\#\#\#\#\#\# \\
    \#\#\# Instruction \\
Using the articles about \{topic keyword\} above, please create a concise timeline with \{$l$\} events following the constraint below. Using only the information from the articles, provide the date and a \{$k$\}-sentence summary for each important event. \\
\\
\#\#\# Constraint \\
\{constraint\} \\
\\
\#\#\# Format \\
YYYY-MM-DD: One-sentence Summary \\
YYYY-MM-DD: One-sentence Summary \\
 \\
\#\#\# Answer \\
{} \\
    \hline \\
\end{tabular}
\caption{Prompt template for the baseline method.}
\label{tab:prompt_baseline}
\end{table}

\end{document}